%% file: main.tex
\pdfoutput=1
\documentclass[11pt]{article}
\usepackage{naacl2021}
\usepackage{enumitem}
\usepackage{times}
\usepackage{latexsym}
\usepackage[T1]{fontenc}
\usepackage[utf8]{inputenc}
\usepackage[utf8]{inputenc} 
\usepackage[T1]{fontenc}    
\usepackage{url}            
\usepackage{booktabs}       
\usepackage{amsfonts}       
\usepackage{nicefrac}       
\usepackage{csquotes}
\usepackage{wrapfig}
\usepackage{xspace}
\usepackage{graphicx}
\usepackage{caption}
\usepackage{subcaption} 
\usepackage[ruled, vlined]{algorithm2e}
\usepackage{array}
\usepackage{makecell}

\newcommand*{\rowstyle}[1]{
	\gdef\@rowstyle{#1}
	\@rowstyle\ignorespaces
}
\newcolumntype{=}{
	>{\gdef\@rowstyle{}}
}
\newcolumntype{+}{
	>{\@rowstyle}
}

\input{math_definition}

\newcommand{\ie}{i.e.\xspace}
\newcommand{\eg}{e.g.\xspace}

\usepackage{microtype}

\title{Revisiting Document Representations for Large-Scale Zero-Shot Learning}

\author{Jihyung Kil \\
  {\normalsize The Ohio State University}\\
  {\normalsize Columbus, Ohio, USA} \\
  {\normalsize \texttt{kil.5@osu.edu}} \And
  Wei-Lun Chao\\
  {\normalsize The Ohio State University}\\
  {\normalsize Columbus, Ohio, USA} \\
  {\normalsize \texttt{chao.209@osu.edu}}
  }
  
\begin{document}
\maketitle
\input{abstract}
\input{intro}
\input{related}
\input{approach}
\input{exp}
\input{disc}
\section*{Acknowledgment}
This research is supported by the OSU GI Development funds.
We are thankful for the support of computational resources by Ohio Supercomputer Center and AWS Cloud Credits for Research.

\bibliographystyle{acl_natbib}
\bibliography{main}
\input{supp_content}

\end{document}

%% file: math_definition.tex
\usepackage{amssymb}
\usepackage{amsmath,amsfonts}
\usepackage{amsopn}
\usepackage{bm} 
\usepackage{multirow}

\newcommand{\vct}[1]{\boldsymbol{#1}} 
\newcommand{\mat}[1]{\boldsymbol{#1}} 

\newcommand{\field}[1]{\mathbb{#1}}
\newcommand{\R}{\field{R}} 

\newcommand{\ProbOpr}[1]{\mathbb{#1}}

\newcommand{\expect}[2]{%
\ifthenelse{\equal{#2}{}}{\ProbOpr{E}_{#1}}
{\ifthenelse{\equal{#1}{}}{\ProbOpr{E}\left[#2\right]}{\ProbOpr{E}_{#1}\left[#2\right]}}} 
\newcommand{\var}[2]{%
\ifthenelse{\equal{#2}{}}{\ProbOpr{VAR}_{#1}}
{\ifthenelse{\equal{#1}{}}{\ProbOpr{VAR}\left[#2\right]}{\ProbOpr{VAR}_{#1}\left[#2\right]}}} 

\DeclareMathOperator{\argmax}{arg\,max}

\newcommand{\vtheta}{\vct{\theta}}

\newcommand{\vx}{{\vct{x}}}

\newcommand{\va}{\vct{a}}

\newcommand{\vh}{\vct{h}}

\newcommand{\vphi}{\vct{\phi}}
\newcommand{\vpsi}{\vct{\psi}}

\newcommand{\mM}{\mat{M}}

\newcommand{\eat}[1]{}

%% file: abstract.tex
\begin{abstract}
Zero-shot learning aims to recognize unseen objects using their semantic representations. Most existing works use visual attributes labeled by humans, not suitable for large-scale applications. In this paper, we revisit the use of documents as semantic representations. We argue that documents like Wikipedia pages contain rich visual information, which however can easily be buried by the vast amount of non-visual sentences. To address this issue, we propose a semi-automatic mechanism for visual sentence extraction that leverages the document section headers and the clustering structure of visual sentences. The extracted visual sentences, after a novel weighting scheme to distinguish similar classes, essentially form semantic representations like visual attributes but need much less human effort. On the ImageNet dataset with over 10,000 unseen classes, our representations lead to a $64\%$ relative improvement against the commonly used ones.
\end{abstract}

%% file: intro.tex
\section{Introduction}
\label{s_intro}

Algorithms for visual recognition usually require hundreds of labeled images to learn how to classify an object~\cite{he2016deep}. In reality, however, the frequency of observing an object follows a long-tailed distribution~\cite{zhu2014capturing}: many objects do not appear frequently enough for us to collect sufficient images. Zero-shot learning (ZSL)~\cite{lampert2009learning}, which aims to build classifiers for unseen object classes using their \emph{semantic representations}, has thus emerged as a promising paradigm for recognizing a large number of classes.

Being the only information of unseen objects, how well the semantic representations describe the visual appearances plays a crucial role in ZSL. One popular choice is \emph{visual attributes}~\cite{lampert2009learning,patterson2012sun,wah2011caltech} carefully annotated by humans. For example, the bird ``Red bellied Woodpecker'' has the ``capped head pattern'' and ``pointed wing shape''. While strictly tied to visual appearances, visual attributes are laborious to collect, limiting their applicability to small-scale problems with hundreds of classes.

\begin{figure}[t]
    \centering
    \includegraphics[width=1\linewidth]{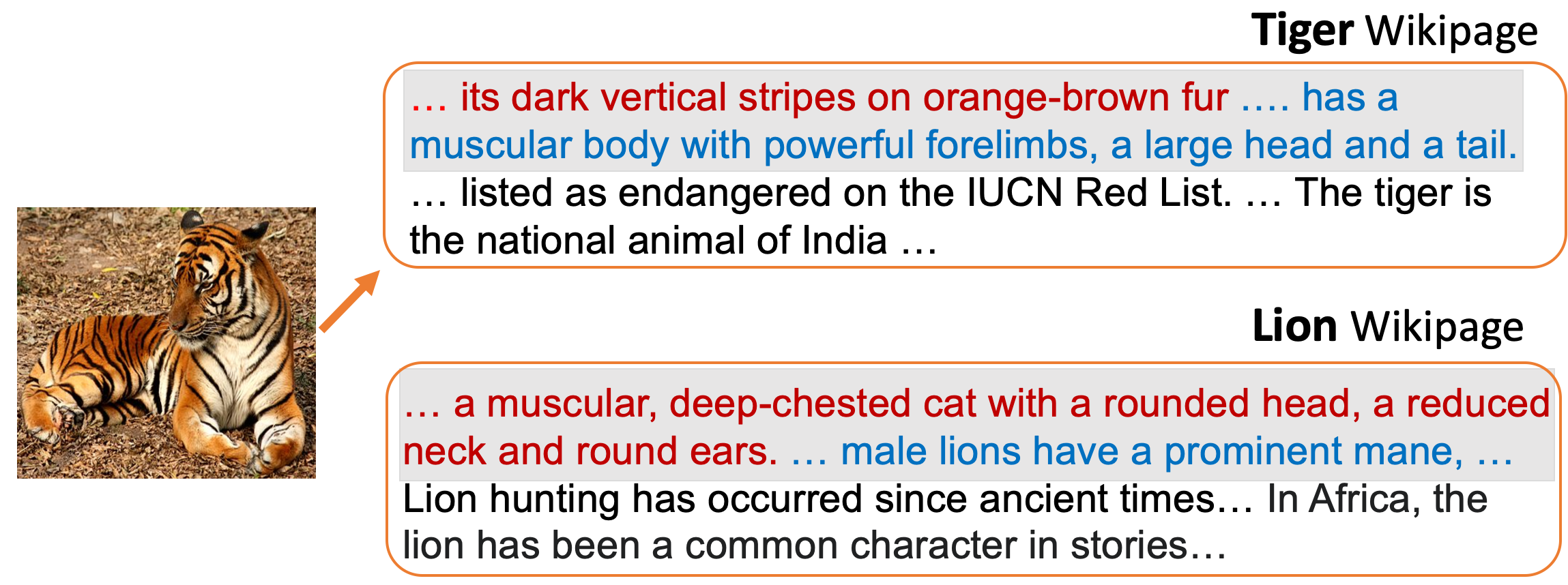}
    \vspace{-20pt}
    \caption{\small An illustration of our ZSL approach, which recognizes the input image by comparing it to the visual sentences of documents. Here we show two documents, one for ``Tiger'' and one for ``Lion''.
    The gray area highlights the extracted visual sentences ({\color{red}red}: by section headers; {\color{blue}blue}: by clustering).
    }
    \label{fig:main}
    \vspace{-10pt}
\end{figure}

For large-scale problems like ImageNet~\cite{deng2009imagenet} that has more than $20,000$ classes, existing ZSL algorithms~\cite{frome2013devise,norouzi2013zero} mostly resort to \emph{word vectors} of classes names~\cite{mikolov2013distributed,pennington2014glove} that are automatically extracted from large corpora like Common Crawl. While almost labor free, word vectors are purely text-driven and barely aligned with visual information. As a result, the state-of-the-art ZSL accuracy on ImageNet falls far behind being practical~\cite{changpinyo2020classifier}.

\emph{Is it possible to develop semantic representations that are as powerful as visual attributes without significant human effort?} A feasibility study by representing a class with its Wikipedia page shows some positive signs --- Wikipedia pages do capture rich attribute information. For example, the page ``Red-bellied Woodpecker'' contains phrases ``red cap going from the bill to the nape'' and ``black and white barred patterns on their back, wings and tail'' that exactly match the visual attributes mentioned above. In other words, if we can identify \emph{visual} sentences from a document to represent a class, we are likely to attain much higher ZSL accuracy\footnote{Representing a class by a document has been studied in~\cite{zhu2018generative,elhoseiny2013write,qiao2016less}, but they use all sentences instead of extracting the visual ones.}.

To this end, we present a simple yet effective semi-automatic approach for \emph{visual sentence extraction}, which leverages two informative semantic cues. First, we leverage the \emph{section structures} of Wikipedia pages: the section header indicates what kind of sentences (visual or not) appear in the section.
Concretely, we search Wikipedia pages of common objects following the sysnsets in ImageNet (\eg, fish, room), and manually identify sections that contain visual information (\eg, characteristics, appearance).
We then apply these visual headers to the Wikipedia pages of the remaining ImageNet classes.
Second, we observe that visual sentences share some common contextual patterns: for example, they contain commonly used words or phrases of visual attributes (\eg, red color, furry surface). To leverage these patterns, we perform K-means sentence clustering using the BERT features~\cite{devlin2018bert} and manually select clusters that contain visual information. We keep sentences in these clusters and combine them with those selected by section headers to represent a document. See~\autoref{fig:main} for an illustration.

To further increase the discriminative ability of the visual sentences between similar object classes (\eg, breeds of dogs), we  introduce a novel scheme to assign weights to sentences, emphasizing those that are more representative for each class.

We validate our approach on three datasets: ImageNet Fall 2011 dataset~\cite{deng2009imagenet}, which contains $14,840$ unseen classes with Wikipedia pages; Animals with Attributes 2 (AwA2)~\cite{xian2018zero}, which has $50$ animal classes; Attribute Pascal and Yahoo (aPY)~\cite{farhadi2009describing}, which has 32 classes.
Our results are promising: compared to word vectors on ImageNet, we improve by {$64\%$} using visual sentences. On AwA2 and aPY, compared to visual attributes annotated by humans, we improve by $8\%$ and $5\%$, respectively. Moreover, our new semantic representations can be easily incorporated into any ZSL algorithms. Our code and data will be available at \url{https://github.com/heendung/vs-zsl}.

%% file: related.tex
\section{Related Work}
\label{s_related}
\noindent\textbf{Semantic representations.} Visual attributes are the most popular semantic representations \cite{lampert2009learning,patterson2012sun,wah2011caltech,zhao2019large}. However, due to the need of human annotation, the largest dataset has only $717$ classes. \citet{reed2016generative,reed2016learning} collect visual sentences for each image, which is not scalable. For large-scale recognition, word vectors~\cite{mikolov2013distributed} have been widely used.
\citet{lu2015unsupervised,kampffmeyer2019rethinking,wang2018zero} explore the use of WordNet hierarchy~\cite{miller1995wordnet}, which may not be available in other applications.

Similar to ours, \citet{akata2015evaluation,elhoseiny2013write,qiao2016less,zhu2018generative} represent classes by documents, by counting word frequencies but not extracting visual sentences. \citet{al2017automatic} extract \emph{single} word attributes, which are not discriminative enough (\eg, ``red cap'' becomes ``red'', ``cap''). None of them works on ZSL with over 1,000 classes.

\citet{hessel2018quantifying, le2020webly} collect images and tags of a class and derives its semantic representation from tags, which is not feasible for unseen classes on ZSL.

\noindent\textbf{Zero-shot learning algorithms.} 
The most popular way is to learn an embedding space in which visual features and semantic representations are aligned and nearest neighbor classifiers can be applied~\cite{changpinyo2017predicting,romera2015embarrassingly,akata2015label,kodirov2017semantic,schonfeld2019generalized,zhu2019generalized,Xie_2019_CVPR,socher2013zero}.
These algorithms consistently improve accuracy on datasets with attributes. Their accuracy on ImageNet, however, is saturated, mainly due to the poor quality of semantic representations~\cite{changpinyo2020classifier}.

%% file: approach.tex
\section{\emph{Visual} Sentence Extraction}
\label{s_approach}

\subsection{Background and notation}\label{background_not} ZSL algorithms learn to align visual features and semantic representations using a set of \emph{seen} classes $\mathcal{S}$. The alignment is then applied to the test images of unseen classes $\mathcal{U}$. We denote by $\mathcal{D} = \{(\vx_n, y_n\in\mathcal{S})\}_{n=1}^N$ the training data (\ie, image feature and label pairs) with the labels coming from $\mathcal{S}$. 

Suppose that we have access to a semantic representation $\va_c$ (\eg, word vectors) for each class $c\in\mathcal{S}\cup\mathcal{U}$, one popular algorithm DeViSE~\cite{frome2013devise} proposes the learning objective
\begin{align}
    \sum_n \sum_{c\neq y_n} \max\{0, \Delta & - f_{\vtheta}^\top(\vx_n)\mM g_{\vphi}(\va_{y_n})\nonumber \\
    & + f_{\vtheta}^\top(\vx_n)\mM g_{\vphi}(\va_{c})\}, \label{e_devise}
\end{align}
where $\Delta\geq0$ is a margin. That is, DeViSE tries to learn transformations $f_{\vtheta}$ and $g_{\vphi}$ and a matrix $\mM$ to maximize the visual and semantic alignment of the same classes while minimizing that between classes. We can then classify a test image $\vx$ by
\begin{align}
\argmax_{c\in\mathcal{U}}f_{\vtheta}^\top(\vx)\mM g_{\vphi}(\va_{c}).
\end{align}
Here, we consider that every class $c\in\mathcal{S}\cup\mathcal{U}$ is provided with a document $H_c=\{\vh^{(c)}_1, \cdots, \vh^{(c)}_{|H_c|}\}$ rather than $\va_c$, where $|H_c|$ is the amount of sentences in document $H_c$ and $\vh^{(c)}_j$ is the $j$th sentence, encoded by BERT~\cite{devlin2018bert}. We mainly study DeViSE, but our approach can easily be applied to other ZSL algorithms.

\begin{table}
\centering
\scriptsize 
\begin{tabular}{l}
\multicolumn{1}{c}{\textbf{Section headers}}\\
\hline
\addlinespace[0.1cm]
{Characteristics, Description, Appearance, Habitat, Diet,}\\ {Construction and Mechanics, Materials for utensil,}\\
Design for appliance, Furnishings for room,  Fabrication,\\
Feature for geological formation, Design, Equipment for sport\\
\hline
\addlinespace[0.1cm]
{History, Health, Terminology, Mythology, Conservation, Culture,}\\{ References, External links, Further reading}\\
\hline
\end{tabular}
\vskip-5pt
\caption{\small  Visual (top) \& Non-Visual (bottom) sections.}
\vskip-10pt
\label{visual_sections}
\end{table} 

\subsection{Visual section selection}
We aim to filter out sentences in $H_c$ that are not describing visual information.
We first leverage the section headers in Wikipedia pages, which indicate what types of sentences (visual or not) are in the sections. For example, the page ``Lion'' has sections ``Description'' and ``Colour variation'' that are likely for visual  information, and ``Health'' and ``Cultural significance'' that are for non-visual information.

To efficiently identify these section headers, we use ImageNet synsets \cite{deng2009imagenet}, which group objects into {$16$} broad categories.
We randomly sample $30\sim35$ classes per group, resulting in a set of $500$ classes. 
We then retrieve the corresponding Wikipedia pages by their names and manually identify section headers related to visual sentences. By sub-sampling classes in this way, we can quickly find section headers that are applicable to other classes within the same groups.
\autoref{visual_sections} shows some visual/non-visual sections gathered from the 500 classes. For example, ``Characteristics'' frequently appears in pages of animals to describe their appearances. In contrast, sections like ``History'' or ``Mythology'' do not contain visual information. Investigating all the $500$ Wikipedia pages carefully, we find $40$ distinct visual sections. We also include the first paragraph of a Wikipedia page, which often contains visual information. 

\subsection{Visual cluster selection}
Our second approach uses K-means for sentence clustering: visual sentences often share common words and phrases of visual attributes, naturally forming clusters.
We represent each sentence using the BERT features \cite{devlin2018bert}, and perform K-means (with $K=100$) over all the sentences from Wikipedia pages of ImageNet classes.
We then manually check the 100 clusters and identify $40$ visual clusters.
\autoref{visual_clusters} shows a visual (top) and a non-visual (bottom) cluster. We highlight sentences related to two classes: ``kit-fox'' (red) and ``tiger'' (blue). The visual cluster describes the animals' general appearances, especially about visual attributes ``dark'', ``black'', ``tail'', ``large'', etc.
In contrast, the non-visual cluster describes mating and lifespan that are not related to visual aspects.

\begin{table}
\centering
\scriptsize
\begin{tabular}{l}
\multicolumn{1}{c}{\textbf{Sentence clusters}}\\
\hline
\addlinespace[0.1cm]
{ \textcolor{red}{It has large ears that help the fox lower its body temperature.}}\\
{ \textcolor{red}{It usually has a gray coat, with rusty tones, and a black tip to its tail. }}\\
{ \textcolor{red}{It has distinct dark patches around the nose.}}\\
{ \textcolor{blue}{It is most recognisable for its dark vertical stripes on orangish-brown fur.}}\\
{ \textcolor{blue}{$\cdots$ muscular body with powerful forelimbs, a large head and a tail.
}}\\
{ \textcolor{blue}{They have a mane-like heavy growth of fur around the neck and jaws $\cdots$}}\\ \hline
\addlinespace[0.1cm]
{ \textcolor{red}{The kit fox is a socially monogamous species.}}\\
{ \textcolor{red}{Male and female kit foxes usually establish monogamous mating $\cdots$ }}\\
{ \textcolor{red}{
The average lifespan of a wild kit fox is 5.5 years.}}\\ 
{ \textcolor{blue}{Tiger mates all year round, but most cubs are born between March $\cdots$}}\\ 
{ \textcolor{blue}{
The father generally takes no part in rearing. 
}}\\
{ \textcolor{blue}{The mortality rate of tiger cubs is about 50\% in the first two years.
}}\\

\hline
\end{tabular}
\vskip-5pt
\caption{\small Sentence clusters. The top cluster is \emph{visual} and the bottom one is \emph{non-visual}. The sentences from a class \emph{kit-fox} are in \textcolor{red}{red} and those from a class \emph{tiger} are in \textcolor{blue}{blue}.}
\vskip-10pt
\label{visual_clusters}
\end{table}

\subsection{Semantic representations of documents}
\label{ssec_wa}
After we obtain a filtered document $\hat{H}_c$, which contains sentences of the \emph{visual} sections and clusters, the next step is to represent $\hat{H}_c$ by a vector $\va_c$ so that nearly all the ZSL algorithms can leverage it.

A simple way is \textbf{average}, $\bar{\va}_c=\frac{1}{|\hat{H}_c|}\sum_{\vh\in \hat{H}_c} \vh$, where $\vh$ is the BERT feature. This, however, may not be discriminative enough to differentiate similar classes that share many common descriptions (e.g., dog classes share common phrase like ``a breed of dogs'' and ``having a coat or a tail''). 

We therefore propose to identify informative sentences that can enlarge the difference of $\va_c$ between classes. Concretely, we learn to assign each sentence a weight $\lambda$, such that the resulting \textbf{weighted average} $\va_c=\frac{1}{|\hat{H}_c|}\sum_{\vh\in \hat{H}_c} \lambda(\vh)\times\vh$ can be more distinctive. We model $\lambda(\cdot)\in\R$ by a multi-layer perceptron (MLP) $b_{\vpsi}$
\begin{align}\label{lambda_h_eq}
    \lambda(\vh)=\frac{\exp(b_{\vpsi}(\vh))}{\sum_{{\vh}^\prime\in \hat{H}_c} \exp(b_{\vpsi}({\vh}^\prime))}. 
\end{align}
We learn $b_{\vpsi}$ to meet two criteria. On the one hand, for very similar classes $c$ and $c'$ whose similarity $\cos(\va_c, \va_{c'})$ is larger than a threshold $\tau$, we want $\cos(\va_c, \va_{c'})$ to be smaller than $\tau$ so they can be discriminable. On the other hand, for other pair of less similar classes, we want their similarity to follow the \textbf{average} semantic representation $\bar{\va}_c$\footnote{The purpose of introducing $\lambda(\cdot)$ is to improve $\va_c$ from the {average} representation $\bar{\va}_c$ to differentiate similar classes.}. 

To this end, we initialize $b_{\vpsi}$ such that the initial $\va_c$ is close to $\bar{\va}_c$. We do so by first learning $b_{\vpsi}$ to minimize the following objective
\begin{align}
    \sum_{c\in\mathcal{S}\cup\mathcal{U}} \max\{0, \epsilon - \cos(\va_c, \bar{\va}_c)\}.
\end{align}
We set $\epsilon=0.9$, forcing $\va_c$ and $\bar{\va}_c$ of the same class to have $\cos(\va_c, \bar{\va}_c)>0.9$. We then fine-tune $b_{\vpsi}$ by minimizing the following objective
\begin{align}\label{less_similar_eq}
    \sum_{c}^{\mathcal{S}\cup\mathcal{U}} \sum_{c \neq c'}^{\mathcal{S}\cup\mathcal{U}} \max\{0, \cos(\va_c, \va_{c'}) - \tau\}.
\end{align}
We assign $\tau$ a high value (\eg, $0.95$) to only penalize overly similar semantic representations.
Please see the appendix for details.

\noindent\textbf{Comparison.} Our approach is different from DAN~\cite{iyyer2015deep}. First, we learn an MLP to assign weights to sentences so that their embeddings can be combined appropriately to differentiate classes. In contrast, DAN computes the averaged embedding and learns an MLP to map it to another (more discriminative) embedding space. Second, DAN leans the MLP with a classification loss. In contrast, we learn the MLP to reduce the embedding similarity between similar classes while maintaining the similarity for other pairs of classes.

%% file: exp.tex
\section{Experiments}
\label{s_exp}

\subsection{Dataset and splits: ImageNet}
\label{ss_data}
We use the ImageNet Fall 2011 dataset \cite{deng2009imagenet} with $21,842$ classes. We use the 1K classes in ILSVRC 2012 \cite{russakovsky2015imagenet} for DeViSE training and validation (cf.~\autoref{e_devise}), leaving the remaining $20,842$ classes as unseen classes for testing. We follow~\cite{changpinyo2016synthesized} to consider three tasks, \text{2-Hop}, \text{3-Hop}, and \text{ALL}, corresponding to \textbf{1,290}, \textbf{5,984}, and \textbf{14,840} unseen classes \emph{that have Wikipedia pages and word vectors} and are within two, three, and arbitrary tree hop distances (w.r.t. the ImageNet hierarchy) to the 1K classes. On average, each page contains \textbf{80} sentences. For images, we use the $2,048$-dimensional ResNet visual features~\cite{he2016deep} provided by~\citet{xian2018zero}. For sentences, we use a $12$-layer pre-trained BERT model~\cite{devlin2018bert}.
We denote by BERT\textsubscript{p} the pre-trained BERT and BERT\textsubscript{f} the one fine-tuned with DeViSE. Please see the appendix for details.

\subsection{Baselines, variants, and metrics}
\label{baseline}
Word vectors of class names are the standard semantic representations for ImageNet.
Here we compare to 
the state-of-the-art \textbf{w2v-v2} provided by \citet{changpinyo2020classifier}, corresponding to a skip-gram model~\cite{mikolov2013distributed} trained with ten passes of the Wikipedia dump corpus. For ours, we compare using all sentences \textbf{(NO)}, visual sections \textbf{(Vis\textsubscript{sec})} or visual clusters \textbf{(Vis\textsubscript{clu})}, and both \textbf{(Vis\textsubscript{sec-clu})}. On average, \textbf{Vis\textsubscript{sec-clu}} filters out \textbf{57$\%$} of the sentences per class. We denote \textbf{weighted average} (Section~\ref{ssec_wa}) by BERT\textsubscript{p-\textbf{w}} and BERT\textsubscript{f-\textbf{w}}. 

The original DeViSE~\cite{frome2013devise} has $f_{\vtheta}$ and $g_{\vphi}$ as identity functions. Here, we consider a stronger version, DeViSE$^\star$, in which we model $f_{\vtheta}$ and $g_{\vphi}$ each by a two-hidden layers multi-layer perceptron (MLP). 
We also experiment with two state-of-the-art ZSL algorithms, EXEM~\cite{changpinyo2020classifier} and HVE~\cite{liu2020hyperbolic}.

We use the average \emph{per-class} Top-1 classification accuracy as the metric~\cite{xian2018zero}.

\begin{table}
\small
\centering
\tabcolsep 4.5pt
\renewcommand{\arraystretch}{1.1}
\begin{tabular}{l|lcccc}
\hline
{Model} & {Type} & {Filter} & {2-Hop} & {3-Hop} & {ALL}\\
\hline
{Random} &  - & - & 0.078 & 0.017 & 0.007 \\
\hline
\multirow{2}{*}{DeViSE} & w2v-v2 & - & 6.45 & 1.99 & 0.78\\
\cline{2-6}
& BERT\textsubscript{p} & No & 6.73 & 2.23 & 0.83\\

\hline
\multirow{9}{*}{DeViSE$^\star$} & w2v-v2 & - & 11.55 & 3.07 & 1.48\\
\cline{2-6}

& & No & 13.84 & 4.05 & 1.75\\
& BERT\textsubscript{p} & Vis\textsubscript{sec} & 15.56 & 4.41 & 1.82\\
& & Vis\textsubscript{clu} & 15.72 & 4.49 & 2.01\\
& & Vis\textsubscript{sec-clu} & 15.86 & 4.65 & 2.05\\
\cline{2-6}
& BERT\textsubscript{p-\textbf{w}} & Vis\textsubscript{sec-clu} & 16.32 & 4.73 & 2.10\\
\cline{2-6}

& & No & 17.70 & 5.17 & 2.29\\
& BERT\textsubscript{f} & Vis\textsubscript{sec} & 19.52 & 5.20 & 2.32\\
& & Vis\textsubscript{clu} & {19.74} & {5.37} & {2.36}\\
& & Vis\textsubscript{sec-clu} & 19.82 & 5.39 & 2.39\\ 
\cline{2-6}
& BERT\textsubscript{f-\textbf{w}} & Vis\textsubscript{sec-clu} & \textcolor{blue}{20.47} & \textcolor{red}{5.53} & \textcolor{red}{2.42}\\ \hline
\multirow{2}{*}{EXEM} & w2v-v2 & - & 16.04 & 4.54 & 1.99 \\ \cline{2-6}
 & BERT\textsubscript{f} & Vis\textsubscript{sec-clu} & \textcolor{red}{21.22} & \textcolor{blue}{5.42} & \textcolor{blue}{2.37} \\ \hline
\multirow{2}{*}{HVE} & w2v-v2 & - & 8.63 & 2.38 & 1.09 \\ \cline{2-6}
 & BERT\textsubscript{f-\textbf{w}} & Vis\textsubscript{sec-clu} & 18.42 & 5.12 & 2.07 \\ \hline
\end{tabular}
\vskip-5pt
\caption{\small Comparison of different semantic representations on ImageNet. We use \emph{per-class} Top-1 accuracy(\%). The best is in \textcolor{red}{red} and the second best in \textcolor{blue}{blue}.
}
\vskip-10pt
\label{results}
\end{table}

\begin{table*}[t!]
\small
\centering
\renewcommand{\arraystretch}{1.1}
\begin{tabular}{l|l|cccc|cccc}
\hline
\multirow{3}{*}{Model} & \multirow{3}{*}{Type} & & & {AwA2} & & & & {aPY} & \\
\cline{3-10}
& & \multirow{2}{*}{ZSL} & & {GZSL} & & \multirow{2}{*}{ZSL} & & {GZSL} &\\
\cline{4-6} \cline{8-10}
& & & {U} & {S} & {H} & & {U} & {S} & {H}\\
\hline
\multirow{3}{*}{DeViSE} & Visual attributes & \textcolor{blue}{59.70} & \textcolor{blue}{17.10} & \textcolor{red}{74.70} & \textcolor{blue}{27.80} & \textcolor{blue}{37.02} & \textcolor{blue}{3.54} & \textcolor{blue}{78.41} & \textcolor{blue}{6.73} \\
\cline{2-10}
& w2v-v2 & 39.56 & 2.18 & 69.29 & 4.22 & 27.67 & 1.68 & \textcolor{red}{85.53} & 3.22 \\
\cline{2-10}
& BERT\textsubscript{p} + Vis\textsubscript{sec-clu} & \textcolor{red}{64.32} & \textcolor{red}{19.79} & \textcolor{blue}{72.46} & \textcolor{red}{31.09} & \textcolor{red}{38.79} & \textcolor{red}{3.94} & 71.60 & \textcolor{red}{7.51} \\
\hline
\end{tabular}
\vskip-5pt
\caption{\small Results on AwA2 and aPY. We compare different semantic representations. Visual attributes are annotated by humans. \textbf{GZSL} is the generalized ZSL setting~\cite{xian2018zero}. In GZSL, \textbf{U}, \textbf{S}, \textbf{H} denote unseen class accuracy, seen class accuracy, and their harmonic mean, respectively. We use \emph{per-class} Top-1 accuracy (\%).
}
\vskip-10pt
\label{AwA2_aPY_results}
\end{table*}

\subsection{Main results}
\label{ss_results} 
\autoref{results} summarizes the results on ImageNet. 
In combining with each ZSL algorithm, our semantic representations \textbf{Vis\textsubscript{sec-clu}} that uses visual sections and visual clusters for sentence extraction outperforms \textbf{w2v-v2}. More discussions are as follows.

\noindent\textbf{BERT vs. \textbf{w2v-v2}.}
For both DeViSE$^\star$ and DeViSE, BERT\textsubscript{p} by averaging all the sentences in a Wikipedia page outperforms {w2v-v2}, suggesting that representing a class by its document is more powerful than its word vector.

\noindent\textbf{DeViSE$^\star$ vs. DeViSE.} Adding MLPs to DeViSE largely improves its accuracy: from $0.78\%$ (DeViSE + {w2v-v2}) to $1.48\%$ (DeViSE$^\star$ + {w2v-v2}) at ALL. In the following, we then focus on DeViSE$^\star$.

\noindent\textbf{Visual sentence extraction.} Comparing different strategies for BERT\textsubscript{p}, we see both \textbf{Vis\textsubscript{clu}} and \textbf{Vis\textsubscript{sec}} largely improves \textbf{NO}, demonstrating the effectiveness of sentence selection. Combining the two sets of sentences (\textbf{Vis\textsubscript{sec-clu}}) leads to a further boost.

\noindent\textbf{Fine-tuning BERT.} BERT can be fine-tuned together with DeViSE$^\star$. The resulting BERT\textsubscript{f} has a notable gain over BERT\textsubscript{p} (\eg, $2.39\%$ vs. $2.05\%$).

\noindent\textbf{Weighted average.} With the weighted average (BERT\textsubscript{p-\textbf{w}}, BERT\textsubscript{f-\textbf{w}}), we obtain the best accuracy.

\noindent\textbf{ZSL algorithms.}  EXEM + {w2v-v2} outperforms DeViSE$^\star$ + {w2v-v2}, but falls behind DeViSE$^\star$ + BERT\textsubscript{p-\textbf{w}} (or BERT\textsubscript{f}, BERT\textsubscript{f-\textbf{w}}).  This suggests that algorithm design and semantic representations are both crucial. Importantly, EXEM and HVE can be improved using our proposed semantic representations, demonstrating the applicability and generalizability of our approach.

\subsection{Results on other datasets}
\label{ss_results_others} 
\autoref{AwA2_aPY_results} summarizes the results on AwA2 \cite{xian2018zero} and aPY \cite{farhadi2009describing}. The former has $40$ seen and $10$ unseen classes; the latter has $20$ seen and $12$ unseen classes. We apply DeViSE together with the $2,048$-dimensional ResNet features~\cite{he2016deep} provided by~\citet{xian2018zero}.
Our proposed semantic representations (\ie, \textbf{BERT\textsubscript{p}} + Vis\textsubscript{sec-clu}) outperform \textbf{w2-v2} and the manually annotated visual attributes on both the ZSL and generalized ZSL (GZSL) settings. Please see the appendix for the detailed experimental setup.
These improved results on ImageNet, AwA2, and aPY demonstrate our proposed method's applicability to multiple datasets.

\begin{table}
\small
\centering
\tabcolsep 3.5pt
\vskip 5pt
\renewcommand{\arraystretch}{1.1}
\begin{tabular}{l|lcccc}
\hline
{Model} & {Type} & {Filter} & 2-Hop & {3-Hop} & {ALL} \\
\hline
\multirow{9}{*} & BERT\textsubscript{p} & No & 13.84 & 4.05 & 1.75 \\
\cline{2-6}
& BERT\textsubscript{p-\textbf{w}-\textbf{direct}} & No & 14.85 & 4.25 & 1.79 \\
\cline{2-6}
& & Par\textsubscript{1st} & 13.48 & 4.10 & 1.78 \\
{DeViSE$^\star$} & & Cls\textsubscript{name} & 14.82 & 3.31 & 1.40 \\
& BERT\textsubscript{p} & Vis\textsubscript{sec} & 15.56 & 4.41 & 1.82 \\
& & Vis\textsubscript{clu} & 15.72 & 4.49 & 2.01 \\
& & Vis\textsubscript{sec-clu} & \textcolor{blue}{15.86} & \textcolor{blue}{4.65} & \textcolor{blue}{2.05} \\
\cline{2-6}
& BERT\textsubscript{p-\textbf{w}} & Vis\textsubscript{sec-clu} & \textcolor{red}{16.32} & \textcolor{red}{4.73} & \textcolor{red}{2.10} \\
\hline
\end{tabular}\
\vskip-5pt
\caption{\small The effectiveness of our visual sentence extraction. \textbf{BERT\textsubscript{p-\textbf{w}-\textbf{direct}}} directly learns visual sentences without our sentence selection. \textbf{Par\textsubscript{1st}} and \textbf{Cls\textsubscript{name}} use the first paragraph and sentences containing the class name, respectively.
}
\vskip-10pt
\label{results_2}
\end{table}

\subsection{Analysis on ImageNet}
\label{ss_further_analysis} 
To further justify the effectiveness of our approach, we compare to additional baselines in \autoref{results_2}.
\begin{itemize} [leftmargin=*, itemsep=-2pt, topsep=2pt]
    \item \textbf{BERT\textsubscript{p-\textbf{w-direct}}}: it directly learns $b_\psi$ (\autoref{lambda_h_eq}) as part of the DeViSE objective. Namely, we directly learn $b_\psi$ to identify visual sentences, without our proposed selection mechanisms, such that the resulting $\va_c$ optimizes \autoref{e_devise}.
    \item \textbf{Par\textsubscript{1st}}: it uses the first paragraph of a document.
    \item \textbf{Cls\textsubscript{name}}: it uses the sentences of a Wikipedia page that contain the class name.
\end{itemize}
As shown in \autoref{results_2}, our proposed sentence selection mechanisms (\ie, Vis\textsubscript{sec}, Vis\textsubscript{clu}, and Vis\textsubscript{sec-clu}) outperform all the three baselines.

%% file: disc.tex
\section{Conclusion}
ZSL relies heavily on the quality of semantic representations. Most recent work, however, focuses solely on algorithm design, trying to squeeze out the last bit of information from the pre-define, likely poor semantic representations. \citet{changpinyo2020classifier} has shown that existing algorithms are trapped in the plateau of inferior semantic representations. Improving the representations is thus more crucial for ZSL. We investigate this direction and show promising results by extracting \emph{distinctive visual} sentences from documents for representations, which can be easily used by any ZSL algorithms.
\label{s_disc}

%% file: supp_content.tex
\newpage
\appendix
\section*{Appendix}
In this appendix, we provide details omitted in the main text.

\begin{itemize}
    \item \autoref{contrib} : contribution
    \item \autoref{more_related} : more related work (cf. Section~\ref{s_related} in the main text)
    \item \autoref{sup_sec_wiki}: detailed statistics of Wikipedia pages (cf. Section~\ref{ss_data} in the main text)
    \item \autoref{w_avg_sent}: weighted average representations (cf. Section~\ref{ssec_wa} in the main text)
    \item \autoref{metrics}: dataset, metrics, and ZSL algorithms (cf. Section~\ref{baseline} in the main text)
    \item \autoref{sup_sec_imp}: implementation details (cf. Section~\ref{ss_results} in the main text)
     \item \autoref{ablation_study}: ablation study (cf. Section~\ref{ss_results} in the main text)
    \item \autoref{sup_qual} qualitative results (cf. Section~\ref{s_approach} in the main text)
\end{itemize}

\section{Contribution}
\label{contrib}
Our contribution is not merely in the method we developed, but also in the direction we explored. As discussed in Section~\ref{s_disc} of the main paper, most of the efforts in ZSL have focused on algorithm design to associate visual features and pre-defined semantic representations. Yet, it is also important to improve semantic representations. Indeed, one reason that ZSL performs poorly on large-scale datasets is the poor semantic representations~\cite{changpinyo2020classifier}. We therefore chose to investigate this direction by revisiting document representations, with the goal to make our contributions widely applicable. To this end, we deliberately kept our method simple and intuitive, but also provided insights for future work to build upon. Our manual inspection identified important properties of visual sentences like the clustering structure, enabling us to efficiently extract them. We chose to not design new ZSL algorithms but make our semantic representations compatible with existing ones to clearly demonstrate the effectiveness of improving semantic representations.

\section{More Related Work}
\label{more_related}
\noindent\textbf{Zero-shot learning (ZSL) algorithms} construct visual classifiers based on semantic representations. Some recent work applies generative models to generate images or visual features of unseen classes~\cite{xian2019f,xian2018feature,zhu2018generative}, so that conventional supervised learning algorithms can be applied.

\noindent\textbf{Knowledge bases} usually contain triplets of entities and relationships. The entities are usually objects, locations, etc. For ZSL, we need entities to be fine-grained (e.g., ``beaks'') and capture more visual appearances. YAGO~\cite{suchanek2008yago} and DBpedia~\cite{zaveri2013user} leverage Wikipedia infoboxes to construct triplets, which is elegant but not suitable for ZSL since Wikipedia infoboxes contain insufficient visual information. Thus, these datasets and construction methods may not be directly applicable to ZSL. Nevertheless, the underlying methodologies are inspiring and could serve as the basis for future work. The datasets also offer inter-class relationships that are complementary to visual descriptions, and may be useful to establish class relationships in ZSL algorithms like SynC~\cite{changpinyo2016synthesized}.

\section{Statistics of Wikipedia Pages}
\label{sup_sec_wiki}
\begin{figure}[t]
    \centering
    \includegraphics[width=1\linewidth]{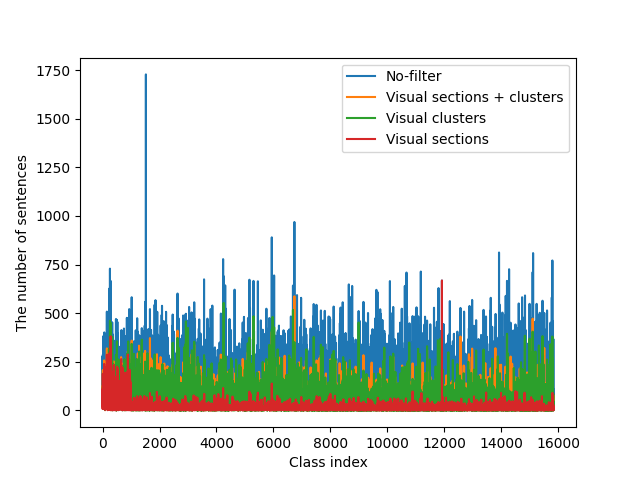}
    \vspace{-20pt}
    \caption{\small Statistics of \textbf{Wikipedia} pages.
    }
    \label{fig:wiki_stats}
\end{figure}

We use a Wikipedia API to extract pages from Wikipedia for ImageNet 21,842 classes. Among 21,842 classes, we find that some classes have multiple Wikipedia pages because of their ambiguous class names. For example, a class ``\emph{black widow}'' in ImageNet refers to a spider with dark brown or a shiny black in colour, but it also refers to the name of a ``\emph{Marvel Comics}'' character in Wikipedia. We therefore exclude such classes and also classes that do not have word vectors, resulting in 15,833 classes. The Wikipedia pages of the 15K classes contain 1,260,889 sentences where each class has 80 sentences on average.
We also investigate the number of sentences by our filters (\ie \textbf{Vis\textsubscript{sec}}, \textbf{Vis\textsubscript{cls}}, \textbf{Vis\textsubscript{sec-clu}}). As a result, we correspondingly find 213,585, 534,852, 542,645 sentences, which are 16\%, 42\%, 43\% of all sentences in 15K classes, respectively (See \autoref{fig:wiki_stats}).

\label{wiki_stats}

\section{Weighted Average Representations}
\label{w_avg_sent}

\subsection{Observation}
Two similar classes may have similar averaged visual sentence embeddings since they share many common descriptions. For example, ~\autoref{fig:kerry_blue} shows that the averaged embedding (\ie, BERT\textsubscript{p} and BERT\textsubscript{f}) between ``Kerry Blue Terrier'' and ``Soft-coated Terrier'' are overly similar since they share a number of sentences containing the common dog features such as ``a breed of dog'' or ``having a coat or a tail''. Thus, if we represent their semantic representations $\va_c$ as the averaged embeddings, ZSL models may not differentiate them. 

\subsection{Algorithm}
In Section~\ref{ssec_wa} of the main text, we introduce $\lambda(\cdot)$ to give each sentence $\vh$ of a document a weight. We note that, while learning $\lambda(\cdot)$ can enlarge the distance of $\va_c$ between similar classes, we should not overly maximize the distance to prevent semantically similar classes (\eg, different breed of dogs) end up being less similar than dissimilar classes (\eg, dogs and cats). To this end, we introduce a margin loss with $\tau$ in ~\autoref{less_similar_eq}, which only penalize overly similar semantic representations.

We also note that, the purpose of $\lambda(\cdot)$ is to improve $\va_c$ from the simple \textbf{average} embedding $\bar{\va}_c$. We therefore initialize $\lambda(\cdot)$ such that the initial $\va_c$ is similar to $\bar{\va}_c$. We do so by first learning $b_{\vpsi}$ with the following objective:

\begin{align}
    \sum_{c\in\mathcal{S}\cup\mathcal{U}} \max\{0, \epsilon - \cos(\va_c, \bar{\va}_c)\}.
\end{align}

We set $\epsilon=0.9$, forcing $\va_c$ and $\bar{\va}_c$ to have a similarity larger than $0.9$.

\subsection{Results}
\begin{figure*}[t]
    \centering
    \includegraphics[width=1\linewidth]{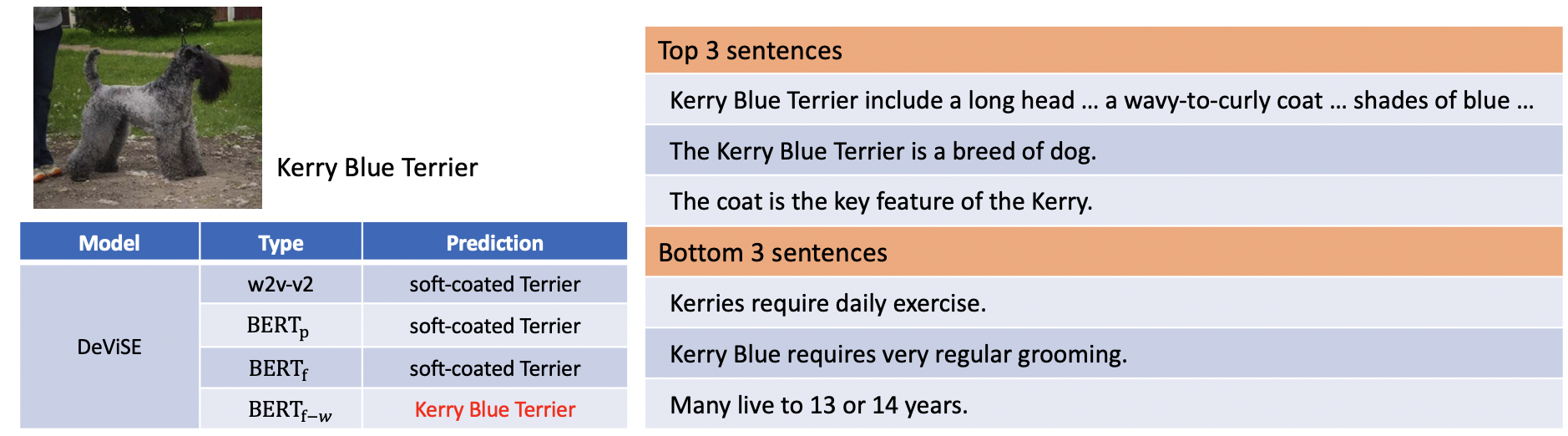}
    \vspace{-20pt}
    \caption{\small Qualitative analysis of a class \emph{Kerry Blue Terrier}. w2v-v2, BERT\textsubscript{p}, and BERT\textsubscript{f} can not distinguish between \emph{Kerry Blue Terrier} and \emph{Soft-coated Terrier} since two classes share the common features of dogs such as ``a breed of dog'' or ``having a coat or a tail''. On the other hand, our weighted average BERT\textsubscript{f-\textbf{w}} is able to differentiate them by weighting on the sentences. We report the Top 3 sentences and the Bottom 3 sentences based on the attention weights.
    }
    \label{fig:kerry_blue}
\end{figure*}

\autoref{fig:kerry_blue} demonstrates the effectiveness of the weighted average embedding BERT\textsubscript{f-\textbf{w}}. While other semantic representations predict ``Kerry Blue Terrier'' as other similar dog, ``soft-coated Terrier'', BERT\textsubscript{f-\textbf{w}} is able to classify the image correctly. In addition, based on the attention weights, we report the Top 3 sentences and the Bottom 3 sentences. The Top 1st sentence contains the inherent features for ``Kerry Blue Terrier'' such as \emph{long head} or \emph{soft-to-curly coat} while the Top 2nd and 3rd sentences describe general features of dogs. On the other hand, the Bottom 3 sentences do not have visual appearance of the object. This suggest that our weighted representation BERT\textsubscript{f-\textbf{w}} is more representative to ``Kerry Blue Terrier'' than other semantic representations.

\section{Dataset, Features, Metrics, and ZSL Algorithm}
\label{metrics}
For visual features, we use the $2,048$-dimensional ResNet visual features~\cite{he2016deep} provided by~\citet{xian2018zero}.
Word vectors can be found in ~\cite{changpinyo2020classifier}.
Followed by~\cite{xian2018zero}, we use the average \emph{per-class} Top-$1$ accuracy as our metric. Instead of simply averaging over all test images (\ie the average \emph{per-sample} Top-$1$ accuracy), this accuracy is obtained by first taking average over all images in each test class independently and then taking average over all test classes. Compared to the average \emph{per-sample} accuracy, the \emph{per-class} accuracy is a more suitable for ImageNet since the dataset is highly imbalanced~\cite{changpinyo2020classifier}. The state-of-the-art algorithms in ZSL are EXEM and HVE proposed by~\cite{changpinyo2020classifier} and~\cite{liu2020hyperbolic}, respectively. To make fair comparison with our models, we evaluate their algorithms on the same number of our test classes using their official codes.

\subsection{ImageNet}
We follow~\cite{xian2018zero,changpinyo2016synthesized} to consider three tasks, \text{2-Hop}, \text{3-Hop}, and \text{ALL}, corresponding to $1,509, 7,678$ and $20,345$ unseen classes that have word vectors and are within two, three, and arbitrary tree hop distances to the $1,000$ seen classes.

We search Wikipedia and successfully retrieve pages for \textbf{15,833} classes, of which \textbf{1,290}, \textbf{5,984}, and \textbf{14,840} are for \text{2-Hop}, \text{3-Hop}, and \text{ALL}.

\subsection{AwA2}
Animals with Attributes2 (AwA2) provides 37,322 images of 50 animal classes. On average, each class includes 746 images. It also provides 85 visual attributes that are manually annotated by humans. In AwA2, classes are split into 40 seen classes and 10 unseen classes. For GZSL, a total of 50 classes is used for testing.

\subsection{aPY}
Attribute Pascal and Yahoo (aPY) contains 15,339 images of 32 classes with 64 attributes. The classes are split into 20 seen classes and 12 unseen classes. A total of 32 classes is used for testing on GZSL.

\subsection{DeViSE~\cite{frome2013devise} vs. EXEM~\cite{changpinyo2020classifier} vs. HVE~\cite{liu2020hyperbolic}}
All algorithms learn feature transformations to associate visual features $\vx$ and semantic representations $\va_c$. The key differences are what and how to learn. DeViSE$^\star$ learns two MLPs $f_{\vtheta}$ and $g_{\vphi}$ to embed $\vx$ and $\va_c$ into a common space, while HVE embeds them into a hyperbolic space. EXEM learns kernel regressors to embed $\va_c$ into the visual space. On how to learn, DeViSE$^\star$ and HVE force each image $\vx$ to be similar to the true class $\va_c$ by a margin loss and a ranking loss respectively, while EXEM learns to regress the averaged visual features of a class from $\va_c$.

\section{Implementation Details}
\label{sup_sec_imp}

\subsection{Sentence representations from BERT} 
Sentence representations can be defined in multiple ways such as a [CLS] token embbedding or an average word embedding from different layers in BERT~\cite{reimers2019sentence}. In our experiments, the average word embedding from the second last layer of BERT achieve the best results in all cases.

\subsection{Hyperparameters}
DeViSE~\cite{frome2013devise} has a tunable margin $\Delta\geq0$ (cf. Section~\ref{background_not} in the main text) which its default value is $0.1$. We try multiple values $0.1$, $0.2$, $0.5$, and $0.7$ to find the best setting. DeViSE uses Adam optimizer which its learning rate is $1e^{-3}$ by default. We try different possible values, $1e^{-3}$, $5e^{-4}$, $2e^{-4}$, and $1e^{-4}$. Among all 16 possible combination of the margin and learning rate, we find that margin of \boldsymbol{$0.2$} and learning rate of \boldsymbol{$2e^{-4}$} achieve the best results on all our cases. 

\subsection{Fine-tuned models}
For fine-tuning, DeViSE$^\star$ is first attached to a BERT model. Then, we train the model with jointly fine-tuning BERT parameters based on the DeViSE$^\star$ objective. Regards to BERT training, ~\citet{houlsby2019parameter} demonstrates that fine-tuning only last few $n$ layers (\eg $2$ or $4$) can outperform fine-tuning all layers in some NLP tasks. ~\citet{kovaleva2019revealing} also shows that the fine-tuning procedure is more effective to the last few layers than earlier layers. Considering the computational resources and time, we therefore set $n$ equal to $2$. After fine-tuning, we freeze BERT parameters and further train DeViSE$^\star$.

\begin{table}
\small
\centering
\tabcolsep 3.5pt
\renewcommand{\arraystretch}{1.1}
\begin{tabular}{l|cccc}
\hline
{Model} & {Type} & {Filter} & {Threshold $\tau$} & 2-Hop\\
\hline
& & &  0.98 & 15.97 \\
\multirow{6}{*}{DeViSE$^\star$} & BERT\textsubscript{p-\textbf{w}} & Vis\textsubscript{sec-clu} & 0.97 & 16.09 \\
& & & 0.96 & {16.32} \\
& & & 0.95 & {16.13} \\
\cline{2-5}
& & &  0.88 & 20.34 \\
& BERT\textsubscript{f-\textbf{w}} & Vis\textsubscript{sec-clu} & 0.86 & \textcolor{blue}{20.44} \\
& & & 0.82 & 20.33 \\
& & & 0.80 & \textcolor{red}{20.47} \\
\hline
\end{tabular}\
\vskip-5pt
\caption{\small Results of per-class Top-1 accuracy(\%) on 2-Hop with different thresholds \emph{$\tau$} and semantic representation types. The best is in \textcolor{red}{red} and the second best in \textcolor{blue}{blue}.
}
\label{abl_diff_ths}
\end{table}

\section{Ablation Study}
\label{ablation_study}
\autoref{abl_diff_ths} shows the results on 2-Hop with different thresholds \emph{$\tau$} introduced in ~\autoref{less_similar_eq}. We obtain the weighted average BERT\textsubscript{p-\textbf{w}} by taking an input $\vh$ from BERT\textsubscript{p} and learning MLP $b_{\vpsi}$ with different \emph{$\tau$} (similar for BERT\textsubscript{f-\textbf{w}}). Then, we measure 2-Hop accuracy based on BERT\textsubscript{p-\textbf{w}} (or BERT\textsubscript{f-\textbf{w}} ). Note that BERT\textsubscript{p} and BERT\textsubscript{f} have different ranges of \emph{$\tau$}, since BERT\textsubscript{f} already has lower similarity between classes. This is because BERT\textsubscript{f} is trained with images (from seen classes) during fine-tuning, which makes BERT\textsubscript{f} more aligned with visual features and thus is more representative. 
We choose $\tau$ based on the ImageNet validation set of the seen classes.

\autoref{boat} shows that the weighted average embedding BERT\textsubscript{p-\textbf{w}} makes similar classes less similar. Originally, a class ``Sea boat'' has overly similar semantic representations with other type of boats (i.e. BERT\textsubscript{p}). After applying our weighting approach, the classes become less similar (e.g. $0.94$ to $0.91$ between ``Sea boat'' and ``Scow'').

\begin{table}
\small
\centering
\renewcommand{\arraystretch}{1.1}
\begin{tabular}{l|ccc}
\hline
\multirow{2}{*}{Class}& Top3 Similar & \multicolumn{2}{c}{Similarity}  \\
& Classes& BERT\textsubscript{p} & BERT\textsubscript{p-\textbf{w}} \\
\hline
& {Scow} & 0.94 & 0.91 \\
{Sea boat} & {Row boat} & 0.93 & 0.91 \\
& {Canoe} & 0.93 & 0.91 \\
\hline
\end{tabular}
\vskip-5pt
\caption{\small Similarity of Top 3 similar classes with \emph{Sea boat} drops after applying the weighting approach.}
\label{boat}
\end{table}

\section{Qualitative Results}
\label{sup_qual}

\subsection{Visual sections and clusters}
\label{sup_sec_res}
We provide additional illustrations of visual sections and clusters of Section~\ref{s_approach} in the main text.

\begin{figure*}[t]
    \centering
    \includegraphics[width=1\linewidth]{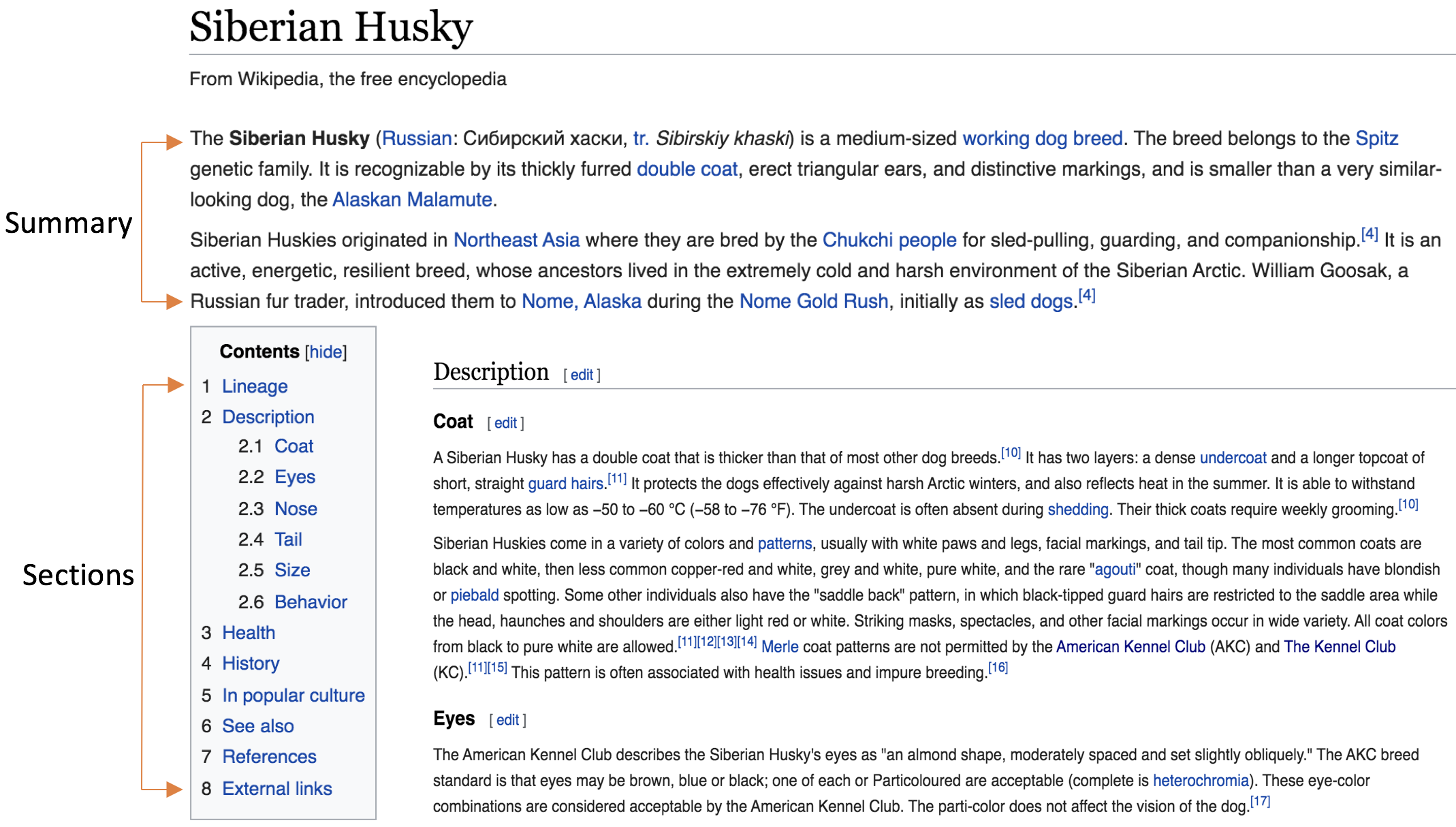}
    \caption{\small Visual sections on \emph{Siberian Husky}.}\label{husky}
\end{figure*}

\autoref{husky} shows visual and non-visual sections in a Wikipedia page \textbf{Siberian Husky}. We note that the summary paragraph and sections such as \emph{Description} contain visual sentences while sections such as \emph{Health} or \emph{History} do not. Similarly, \autoref{visual_clusters_supp} shows two clusters: the top cluster is visual, consisting of information about \emph{hunting} and \emph{preys} of animals while the bottom cluster includes \emph{mythology} sentences not visually related. 

\begin{table}
\centering
\small
\scriptsize
\begin{tabular}{l}
\multicolumn{1}{c}{\textbf{Clusters}}\\
\hline
\addlinespace[0.1cm]
{$\cdots$hunt shortly after sunset, eating small animals $\cdots$}\\
{$\cdots$ if food is scarce, it has been known to eat tomatoes $\cdots$}\\
{Tigers are capable of taking down larger prey like adult gaur  $\cdots$ 
}\\
{Tigers will also prey on such domestic livestock as cattle, horses, $\cdots$
}\\
\hline
\addlinespace[0.1cm]
{Panda is a Roman goddess of peace and travellers $\cdots$}\\
{The Ibex is also a national emblem of the great ancient Axum empire.}\\
{In Aztec mythology, the jaguar was considered to be the totem animal of $\cdots$ }\\ 
{It is the national animal of Guyana, and is featured in its coat of arms $\cdots$}\\ 
\hline
\end{tabular}
\vskip-5pt
\caption{\small K-means sentence clusters. The top cluster has \emph{visual} information about \emph{hunting} and \emph{preys} while the bottom one contains \emph{non-visual} description such as \emph{mythology}.}
\vskip-10pt
\label{visual_clusters_supp}
\end{table}

\subsection{On ImageNet}
\begin{figure*}[t]
    \centering
    \includegraphics[width=1\linewidth]{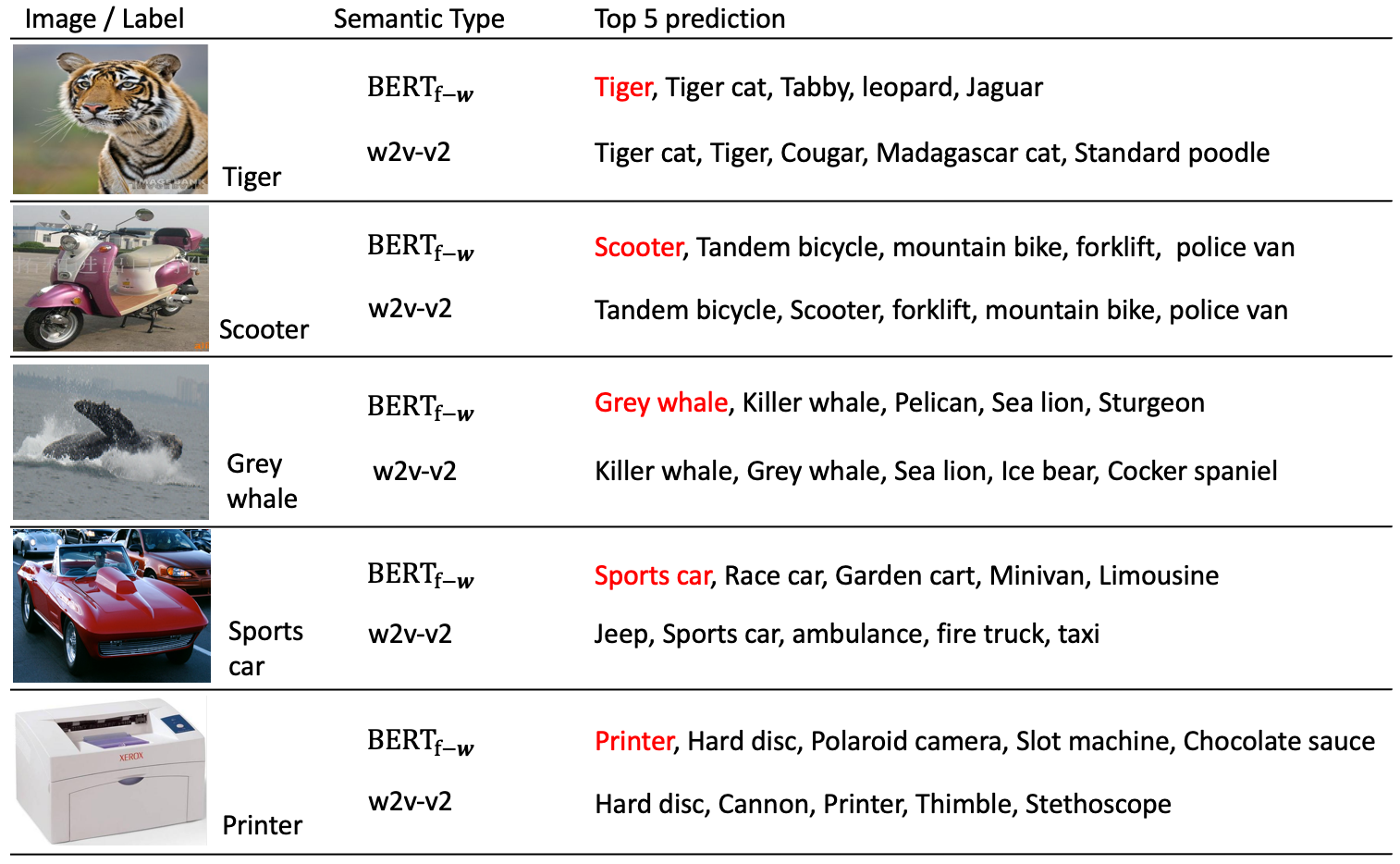}
    \caption{\small Qualitative results between \textbf{BERT\textsubscript{f-\textbf{w}}} and \textbf{w2v-v2} on ImageNet. For each image, we report Top 5 prediction. While \textbf{w2v-v2} is not able to distinguish similar classes (e.g. Predicting ``Scooter'' as ``Tandem bicycle''), our \textbf{BERT\textsubscript{f-\textbf{w}}} differentiates them.}\label{qual_results_sup}
\end{figure*}

\autoref{qual_results_sup} shows the qualitative results of our BERT\textsubscript{f-\textbf{w}} and w2v-v2 on ImageNet. For each image, we provide its label and the Top 5 prediction by BERT\textsubscript{f-\textbf{w}} and w2v-v2. While w2v-v2 is not able to differentiate the similar classes (e.g. Predicting ``Scooter'' as ``Tandem bicycle''), our BERT\textsubscript{f-\textbf{w}} can distinguish them. We also note that the Top 5 classes predicted by BERT\textsubscript{f-\textbf{w}} are similar (e.g. ``Grey whale'' and ``Killer whale''). This suggests that our approach maintains the order of similarity among classes but make their semantic representations more distinctive.

